\documentclass{article}
\usepackage{ijcai23}

\usepackage{times}
\usepackage{soul}
\usepackage{url}
\usepackage[hidelinks]{hyperref}
\usepackage[utf8]{inputenc}
\usepackage[small]{caption}
\usepackage{graphicx, subfigure}
\usepackage{amsmath}
\usepackage{amsthm}
\usepackage{booktabs}
\usepackage{algorithm}
\usepackage{algorithmic}
\usepackage[switch]{lineno}
\usepackage{xcolor}
\usepackage{bm, amsfonts}
\usepackage{cancel}
\definecolor{blue}{HTML}{00B0F0}
\definecolor{green}{HTML}{8DBD6C}
\definecolor{blue2}{HTML}{7CAFDD}
\definecolor{yellow}{HTML}{FFC619}

\usepackage{multirow}
\usepackage{graphicx}

\usepackage{pifont}
\usepackage{newtxtext,newtxmath}
\newcommand{\cmark}{\ding{51}}%
\newcommand{\xmark}{\ding{55}}%

\title{Learning Summary-Worthy Visual Representation for Abstractive \\Summarization in Video}

\author{
	Zenan Xu$^1$\thanks{Work is done during internship at Noah’s Ark Lab.}
	\and
	Xiaojun Meng$^2$\and
	Yasheng Wang$^2$\and
	Qinliang Su$^{1,4}$\thanks{Corresponding author.}\and
	\\Zexuan Qiu$^3$\and
	Xin Jiang$^2$\And
	Qun Liu$^2$
	\affiliations
	$^1$School of Computer Science and Engineering, Sun Yat-sen University, Guangzhou, China\\
	$^2$Noah's Ark Lab, Huawei Technologie\\
	$^3$The Chinese University of Hong Kong, Hong Kong SAR\\
	$^4$Guangdong Key Laboratory of Big Data Analysis and Processing, Guangzhou, China
	\emails
	\{xuzn@mail2, suqliang@mail\}.sysu.edu.cn,
	qzexuan@link.cuhk.edu.hk,\\
	\{xiaojun.meng, wangyasheng, Jiang.Xin, qun.liu\}@huawei.com
}

\begin{document}
	
	\maketitle
	
	\begin{abstract}
		Multimodal abstractive summarization for videos (MAS) requires generating a concise textual summary to describe the highlights of a video according to multimodal resources, in our case, the video content and its transcript. Inspired by the success of the large-scale generative pre-trained language model (GPLM) in generating high-quality textual content (e.g., summary), recent MAS methods have proposed to adapt the GPLM to this task by equipping it with the visual information, which is often obtained through a general-purpose visual feature extractor. However, the generally extracted visual features may overlook some summary-worthy visual information, which impedes model performance. In this work, we propose a novel approach to learning the summary-worthy visual representation that facilitates abstractive summarization. Our method exploits the summary-worthy information from both the cross-modal transcript data and the knowledge that distills from the pseudo summary. Extensive experiments on three public multimodal datasets show that our method outperforms all competing baselines. Furthermore, with the advantages of summary-worthy visual information, our model can have a significant improvement on small datasets or even datasets with limited training data.
	\end{abstract}
	
	\section{Introduction}
	
	With the increasing popularity of video in user-generated content in recent years~\cite{kim2021structured,Cherian2022251DSS}, a large number of open-domain videos have been posted on the Web (e.g., YouTube). Usually, many of the videos are not accompanied by a briefly introduction to reflect the corresponding salient information, which prevents users from quickly finding their interested videos unless taking time to peruse each video. Obviously, in this scenario, it would be valuable to develop an automatic abstractive summarization model~\cite{libovicky2018multimodal} that detects the highlights of each video and then generates a short textual description. 
	
\begin{figure}[!tp]
		\centering
		\includegraphics[scale=0.4]{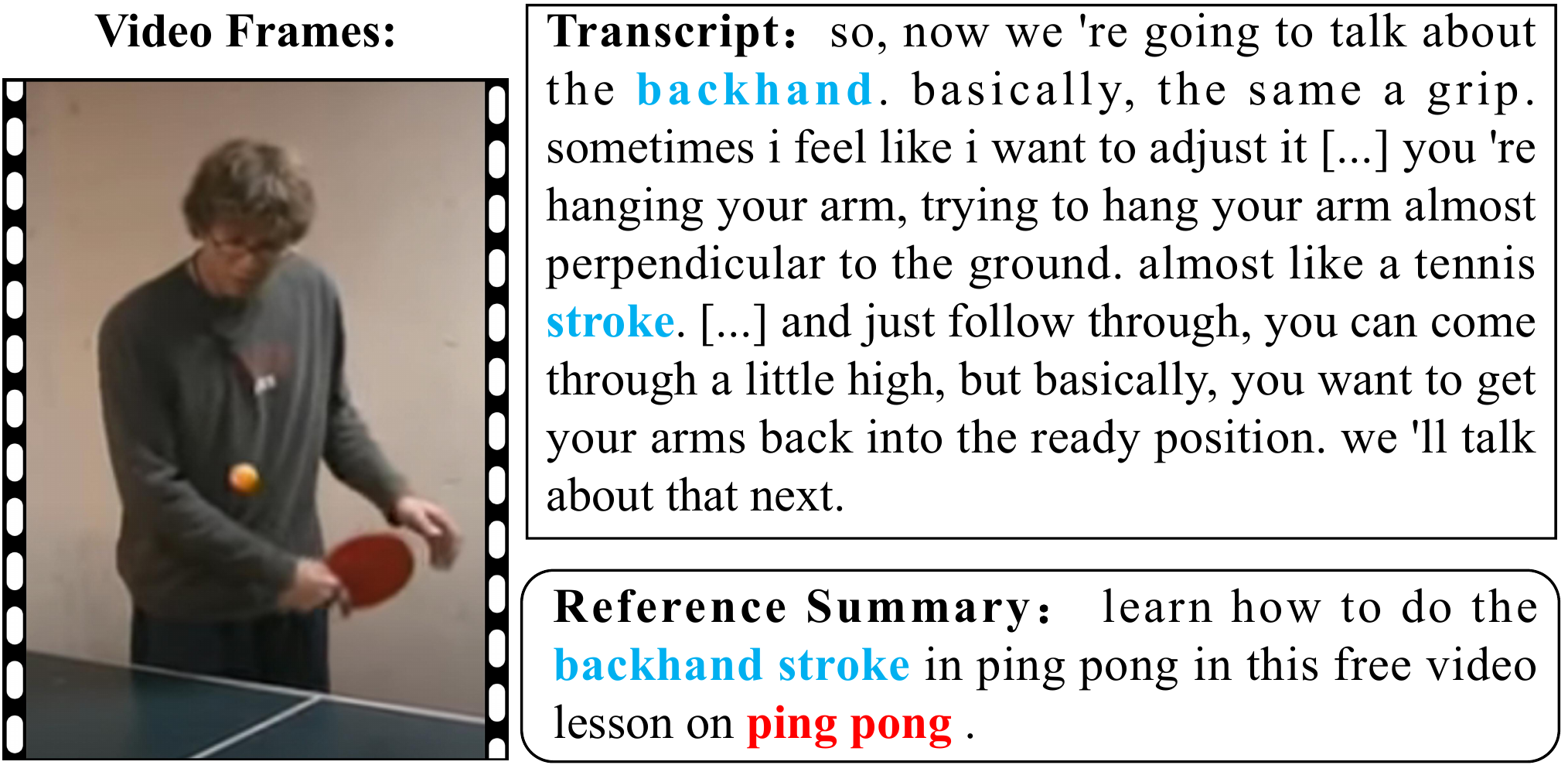}
		\caption{An example of multimodal abstractive summarization. The unimportant textual content is omitted and replaced by [...]. It can be seen that some critical information in the reference is either highlighted in the transcript (e.g., \textbf{\textcolor{blue}{backhand stroke}}) or only available from the video (e.g., \textbf{\textcolor{red}{ping pong}}).}
		\label{fig:introduction}
	\end{figure}

	As illustrated in Figure~\ref{fig:introduction}, the task of multimodal abstractive summarization (MAS) aims to generate a concise textual summary according to multimodal resources, i.e., the video content and its transcript~\cite{sanabria2018how2}. This task is challenging since both visual and textual modalities are complementary to each other, and thus, how to efficiently combine the multimodal information is the key to this task. To leverage information from both modalities, \cite{Palaskar2019MultimodalAS} employed separate encoders on visual and textual data, which was followed by a joint decoder with an attention mechanism to capture the intrinsic connection between the two modalities. Later,  MFFG~\cite{Liu2020MultistageFW} and SFFG~\cite{Liu2022AbstractiveSF} brought the multimodal interaction into the encoder to obtain the fine-grained correlation between multimodal inputs to exploit the complementary information of each modality and achieved promising results. 
	
	\begin{figure*}[!t]
		\centering
		\includegraphics[scale=0.6]{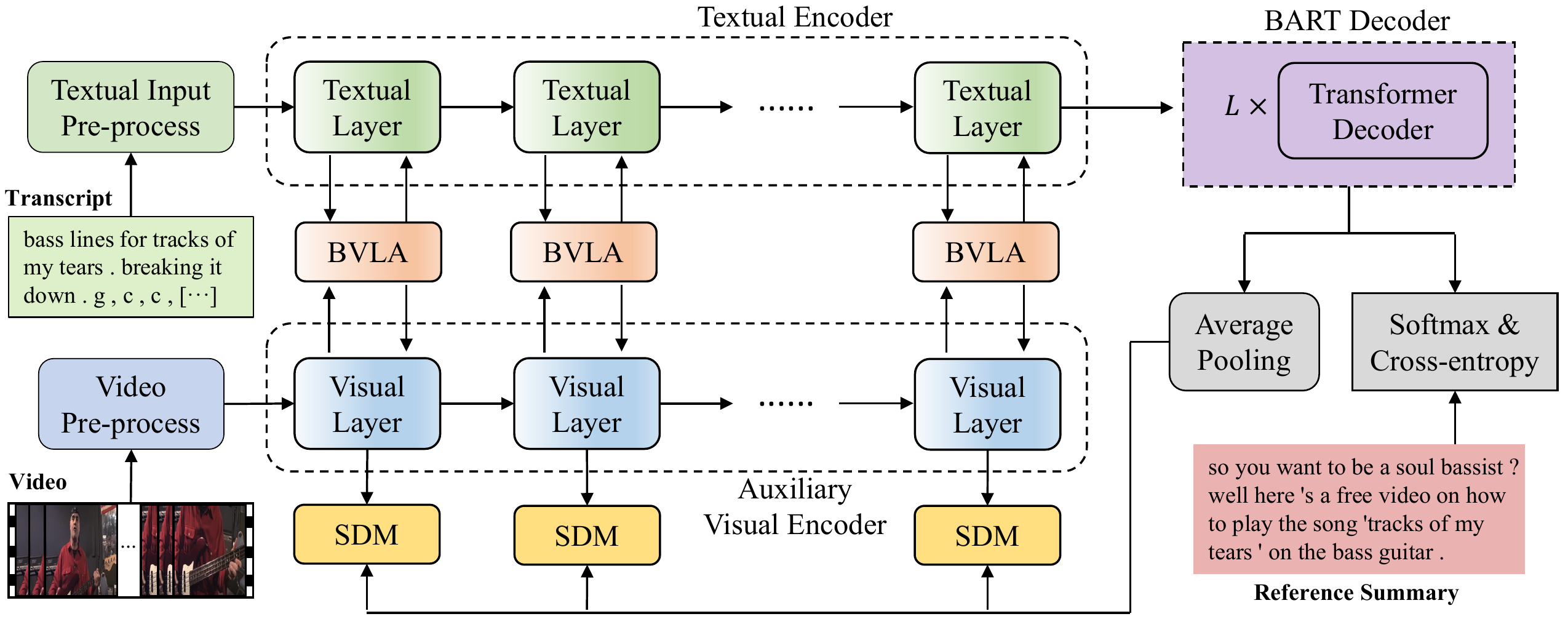}
		\caption{An overview of our proposed $SWVR$ (Summary-Worthy Visual Representation) method. The Bi-directional Visual-Language Attention mechanism (BVLA) and Self Distillation Mechanism (SDM) are introduced to encourage the visual encoder to exploit the summary-wothy information.}
		\label{fig:methodology}
	\end{figure*}
	
	Recently, inspired by the success of the large-scale generative pre-trained language model (GPLM)~\cite{Lewis2019BARTDS,raffel2020exploring} on generating high-quality textual content (e.g., summary),  researchers start to apply it to the MAS task. To do this, VG-GPLM~\cite{yu2021vision} first extracts the video representation with the pre-trained visual feature extractor. Then, to allow the cross-modal interaction, the obtained visual feature of video is injected into each encoder layer of the GPLM with an attention-based text-vision fusion module. This attention module is designed to select the relevant visual information (key \& value) based on the textual features (query). VG-GPLM has demonstrated the potentials of injecting visual representation into language models to improve the MAS performance.
	
	However, existing methods mainly use a general-purpose visual feature extractor,
	which may potentially overlook some summary-worthy visual information, including the essentials that are merely available in the video (e.g., the `\texttt{ping pong}' in Figure~\ref{fig:introduction} only appears in video frames), and thus impede the quality of the generated summary. It is also possible for the existing methods to generate the novel concepts like `\texttt{ping pong}' if sufficient labelled data is provided to well-train the general-purpose extractor to identify task-specific objects. Unfortunately, it may lead to the contradiction between the data hungriness issue of the extractor and the scarcity of annotated summarization data.
	
	To address the above issue, we aim to design a visual encoder that is aware of summary-worthy information. We present a novel method, named $SWVR$, to learn \textbf{S}ummary-\textbf{W}orthy \textbf{V}isual \textbf{R}epresentation for the MAS task. We use a \textbf{B}i-directional \textbf{V}isual-\textbf{L}anguage \textbf{A}ttention mechanism (BVLA) to encourage the visual encoder to exploit the summary-wothy information from the textual data, i.e., transcripts. Afterward, we further introduce a \textbf{S}elf \textbf{D}istillation \textbf{M}echanism (SDM) that takes the generated pseudo summary as the teacher to guide the learning of the visual encoder. This self-distillation encourages the visual encoder to have the ability of capturing and aligning to summary-worthy knowledge that appears in the generated summary after the decoder. Given the surprising and strong text generation ability of GPLM, such distillation helps the visual encoder pay more attention on corresponding key concepts in video frames. Since a sub-optimal placement may impede the model's performance, we enumerate almost all possible ways to combine and evaluate BVLA and SDM modules in the visual encoder. Experimental results on three public datasets show that our method outperforms all competing baselines, especially on the dataset with a relatively small size. Further studies demonstrate the effectiveness of each component, and suggest that the learned summary-worthy feature can help the model identify valuable information, thus benefiting the MAS task.

	\section{Related Work}
	\paragraph{Text-based Abstractive Summarization:}
	Given a long article, abstractive text summarization aims to generate a brief summary that describes the article's most salient content. Early studies were based on statistical or linguistic rules, including extractive and compression~\cite{Knight2002SummarizationBS,Clarke2010DiscourseCF}, templates~\cite{Zhou2004TemplateFilteredHS}, and statistics~\cite{Banko2000HeadlineGB}. Later, the availability of large-scale summarization corpora has promoted the development of various neural networks, among which the representative Seq2Seq model~\cite{Sutskever2014SequenceTS} and the attention mechanism have greatly advanced the quality of summaries ~\cite{paulus2018deep,Wang2019BiSETBS,zhang2020pegasus,Yu2021AdaptSumTL}. Recently, in light of the powerful generative abilities, the large-scale pre-trained language models have led the mainstream in this field~\cite{Lewis2019BARTDS,raffel2020exploring,zhang2020pegasus,qi2020prophetnet}. 
	
	\paragraph{Video-based Abstractive Summarization:}
	Multimodal summarization has been developed for decades~\cite{Erol2003MultimodalSO,Tjondronegoro2011MultimodalSO,Evangelopoulos2013MultimodalSA,Shah2016LeveragingMI,Zhang_Meng_Wang_Jiang_Liu_Yang_2022}. The key method behind it, i.e., multimodal learning, has recently attracted a number of researchers' interest, while in fact little attention has been paid to the video content based summarization. Previous methods mainly focus on a simple situation where the video data contains synchronized signals, e.g., synchronized voice and captions. To tackle the video-based summarization in a more general and asynchronous scenario, \cite{Li2017MultimodalSF} collected a multimodal dataset containing 500 videos of English news articles with human-generated reference summaries. Later, to better promote the development of the MAS for videos, \cite{sanabria2018how2} introduced a large-scale human-annotated video dataset named How2, which contains videos of 2000 hours, and each video is annotated with a short summary. Thanks to the How2 dataset, of which the advent has accelerated the development of MAS methods using neural networks, e.g., the hierarchical attention in \cite{Palaskar2019MultimodalAS}, the forget gate mechanism in  \cite{Liu2020MultistageFW}, and the multi-stage fusion network in \cite{Liu2022AbstractiveSF}. To further leverage the GPLM's generation ability, VG-GPLM~\cite{yu2021vision} studies multiple fusion methods that inject the visual information into GPLMs to improve the MAS for videos. However, VG-GPLM obtains the visual feature via a general-purpose visual encoder, which likely ignores task-specific visual clues that are valuable to summarization. In contrast, our method bridges this gap by learning and then injecting the summary-worthy visual feature into the GPLMs. 
	
	\section{Methodology}
	
	\subsection{Problem Definition}
	Given a video $V$ and its associated textual transcript $T$, our multimodal summarization system is required to summarize the video by a generated concise summary $S$ with maximum probability $p(S|V,T;\theta)$, where $\theta$ stands for the model parameters. It is worth knowing that both the transcript $T$ and generated summary $S$ are in the form of a sequence of words.
	
	For the text-based abstractive summarization task, a prevailing way is to adopt the powerful GPLM, whose ability of generating high-quality texts has been widely demonstrated~\cite{qi2020prophetnet,zhang2020pegasus}. When in the case of multimodal input like a video, a natural idea is to inject the visual features into the encoder of GPLM and then utilize its superior generation to obtain a better textual summary, which is how VG-GPLM~\cite{yu2021vision} works. In the following, we present the proposed $SWVR$ model in detail, and the overall framework is shown in Figure~\ref{fig:methodology}.
	
	\subsection{Video Pre-processing}
	Given a video, we follow previous work~\cite{yu2021vision} to first extract the visual features with 2048 dimension for every 16 non-overlapping frames through a 3D ResNeXt-101 model~\cite{Hara2017CanS3}, which is pre-trained on the Kinetics dataset~\cite{kay2017kinetics}. Then a linear layer is adopted to project the visual features into $d$ dimension vector space and obtain $\bm{Z}_v\in\mathbb{R}^{n_v\times d}$, where $n_v$ is the number of visual tokens.
	
	\subsection{Method Overview}
	As discussed above, we employ the GPLM as our backbone. Specifically, we construct our method with the sequence-to-sequence BART model~\cite{lewis2020bart}, which consists of a textual encoder and a left-to-right decoder designed to generate the textual summary. Given the transcript as input, the textual encoder first tokenizes it and then embeds it into the textual features $\bm{Z}_t\in\mathbb{R}^{n_t\times d}$, which will be fed into the transformer~\cite{vaswani2017attention} encoder layers to exploit the contextual representation as:
	\begin{equation} \label{equ:multiattn}
	\bm{Z}^{l'}_t = LN(MultiAttn(\bm{Z}^{l-1}_t) + \bm{Z}^{l-1}_t),
	\end{equation}
	\begin{equation} \label{equ:ffn}
	\bm{Z}^{l}_t = LN(FFN(\bm{Z}^{l'}_t) + \bm{Z}^{l'}_t),
	\end{equation}
	where $LN$ denotes the layer normalization~\cite{ba2016layer}, $l\in[1,L]$ denotes the $l$-th textual encoder layer, $\bm{Z}^{0}_t$ is initialized with $\bm{Z}_{t}$, and $MultiAttn$ and $FFN$ denotes two sub-layers, i.e., the multi-headed self-attention mechanism and a feed-forward network, respectively\footnote{Additionally, there are residual operations inside each layer, which we omit here for brevity. For the interested readers, we kindly refer to \cite{vaswani2017attention} for more description.}. 
	
	To inject the visual information into each layer of the textual encoder, we add a third sub-layer into each textual encoder layer. Specifically, assume the visual feature that is to be injected into the $l$-th textual encoder layer is $\bm{Z}_v^{l}$, then this sub-layer operates as: 
	\begin{equation} \label{equ:cross-fusion}
	\tilde{\bm{Z}}^{l}_t = LN(VLA(\bm{Z}_v^{l}, \bm{Z}^{l}_{t})\cdot\bm{W}_t^l + \bm{Z}^{l}_{t}),
	\end{equation}
	where $\bm{W}_t^l\in\mathbb{R}^{d\times d}$ is the model weight, and $VLA(\bm{Z}_v^{l}, \bm{Z}^{l}_{t})$ is a unidirectional visual-language attention function that extracts the relevant information from visual feature $\bm{Z}_v^{l}$ according to the textual feature $\bm{Z}^{l}_{t}$, which will be described later. It can be seen from \eqref{equ:cross-fusion} that $\bm{Z}_v^{l}$ takes a great impact on our model. Therefore, our paper focuses on how to obtain a high-quality collection of visual feature, i.e., $\bm{C}_v=\{\bm{Z}_v^{1}, \cdots, \bm{Z}_v^{L}\}$, where each $\bm{Z}_v^{l}$ can be injected into the $l$-th textual encoder layer.
	
	Afterward, the output $\tilde{\bm{Z}}^{L}_{t}$ from the last layer of textual encoder will be provided to the decoder to generate a sequence $(\bm{w}_1, \dots, \bm{w}_{n_w})$ of vectors one element at a time. Since the decoder is auto-regressive, at each step, the generated vector $\bm{w}_i$ is mapped to a new vector of vocabulary size, followed by a softmax function to output the summary word distribution $\hat{\bm{y}}_i$. The negative log-likelihood between the generated summary and the ground-truth summary is used to calculate the loss as:
	\begin{equation}
	\mathcal{L}_{Abs} = -\sum_{j=1}^{n_w} \bm{y}_i\log p(\hat{\bm{y}}_i).
	\end{equation}
	
	\begin{figure}[!t]
		\centering
		\subfigure[Direct utilization of $\bm{Z}_v$.]{
			\begin{minipage}[t]{0.5\linewidth}
				\centering
				\includegraphics[scale=0.55]{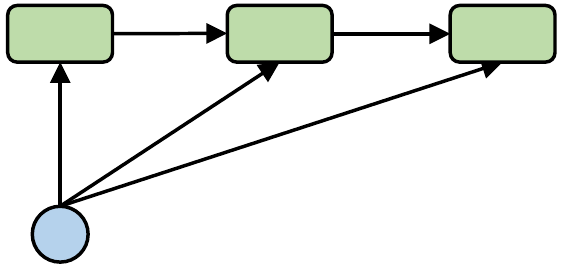}
				\label{fig:methodology_a}
			\end{minipage}
		}%
		\centering
		\subfigure[Auxiliary Visual Encoder.]{
			\begin{minipage}[t]{0.5\linewidth}
				\centering
				\includegraphics[scale=0.55]{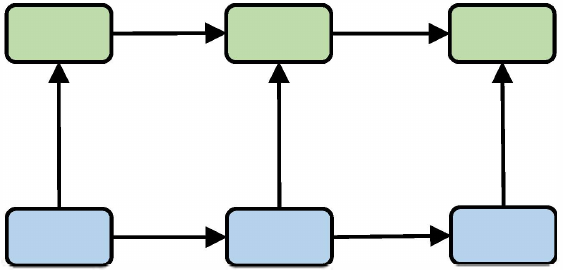}
				\label{fig:methodology_b}
			\end{minipage}%
		}
		\subfigure[Learning SWR from Transcript.]{
			\begin{minipage}[t]{0.5\linewidth}
				\centering
				\includegraphics[scale=0.55]{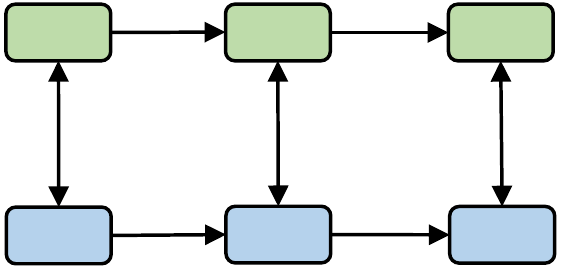}
				\label{fig:methodology_c}
			\end{minipage}
		}%
		\centering
		\subfigure[Learning SWR from Summary.]{
			\begin{minipage}[t]{0.5\linewidth}
				\centering
				\includegraphics[scale=0.55]{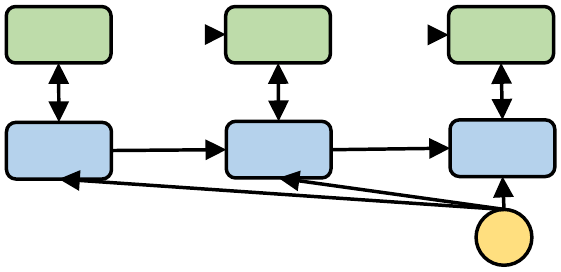}
				\label{fig:methodology_d}
			\end{minipage}%
		}%
		\caption{The simple illustration of different types of visual information injected into the textual encoder layers. SWR stands for summary-worthy representation. The textual and visual information are represented in \bm{\textcolor{green}{green}} and \bm{\textcolor{blue2}{blue}} color, respectively. The \bm{\textcolor{yellow}{yellow}} color stands for the feature map of generated summary.}
		\label{fig:methodology_simple}
	\end{figure}

	\subsection{Learning Summary-Worthy Visual Representation}
	A simple way to generate the collection $\bm{C}_v$ is that we fill $\bm{C}_v$ with the obtained $\bm{Z}_v$ after video pre-processing, namely, $\bm{C}_v=\{\bm{Z}_v^{l} = \bm{Z}_v, \forall l = 1, \cdots, L\}$. In this way, the cross-modal interaction can be simply summarized as Figure~\ref{fig:methodology_a}.
	
	\subsubsection{Auxiliary Visual Encoder}
	However, the direct utilization of $\bm{Z}_v$ may suffer from the below shortcomings: ($i$) The temporal information of a video is ignored, which limits the expressiveness of visual feature; ($ii$) Since recent study~\cite{Raghu2021DoVT} finds that different encoder layers exploit different-level semantic, injecting the same visual feature into different textual encoder layers may potentially restrict the model to learn more meaningful cross-modal interactions.
	
	To address the above issues, also inspired by the recent success of the two-tower architecture~\cite{Radford2021LearningTV,Jia2021ScalingUV}, we introduce an auxiliary visual encoder that takes the video feature as input and learns visual information at different levels of semantics. Specifically, the trainable positional encodings are first added to the $\bm{Z}_v$ to preserve the temporal information. Afterward, this $\bm{Z}_{v}$ can be used directly as input to vanilla transformer-based visual encoder of $L$ layers. 
	Formally, given the input $\bm{Z}_{v}$, the output of each visual layer is computed as:
	\begin{equation}
	\bm{Z}^{l'}_v = LN(MultiAttn(\bm{Z}^{l-1}_v) + \bm{Z}^{l-1}_v),
	\end{equation}
	\begin{equation}
	\bm{Z}^{l}_v = LN(FFN(\bm{Z}^{l'}_v) + \bm{Z}^{l'}_v),
	\end{equation}
	where $\bm{Z}^{0}_v$ is initialized with $\bm{Z}_v$. We collect the output of each visual encoder layer and obtain a new visual collection $\bm{C}_v=\{\bm{Z}_v^{1}, \cdots, \bm{Z}_v^{L}\}$, as shown in Figure~\ref{fig:methodology_b}. In this way, we inject the visual information into the textual feature from the same `height' (e.g., both from the $l$-th layer), which could benefit the cross-modal correlation at different levels of semantics.
	
	\subsubsection{Summary-Worthy Information from Transcript}
	It can be observed that this introduced auxiliary visual encoder is still a general-purpose visual feature extractor. Since not all the information from the visual modality is valuable for summarization~\cite{Liu2020MultistageFW}, the noise and redundancy in the generally learned visual feature makes it less effective for the model to summarize the highlights. We are thus motivated to propose to learn the summary-worthy visual representation. 
	
	Since the summary-worthy information from each modality may be complementary to the other, carrying this observation, we propose to exploit the summary-worthy information from the transcript, which may be complementary to and benefit learning the visual feature. Therefore, we use the multi-head cross-attention to inject the summary-worthy information from the transcript into the visual representation. Technically, we extend the unidirectional $VLA(\cdot, \cdot)$ function (which is employed in \eqref{equ:cross-fusion}) into bi-directional visual-language attention function $BVLA(\cdot, \cdot)$. Given textual feature $\bm{Z}^{l}_t\in\mathbb{R}^{n_t\times d}$ and visual representation $\bm{Z}^{l}_v\in\mathbb{R}^{n_v\times d}$, the $BVLA(\cdot, \cdot)$ will function as:
	\begin{equation}
	[\bm{Z}_{t\rightarrow v}^l, \bm{Z}_{v\rightarrow t}^l] = BVLA(\bm{Z}^{l}_t, \bm{Z}^{l}_v),
	\end{equation}
	where $\bm{Z}_{t\rightarrow v}^l\in\mathbb{R}^{n_v\times d}$ and $\bm{Z}_{v\rightarrow t}^l\in\mathbb{R}^{n_t\times d}$ denote the information of transcript that is relevant to video or vice versa, respectively. Let us take $\bm{Z}_{t\rightarrow v}^l$ as an example to elaborate. Particularly, the $\bm{Z}_{t\rightarrow v}^l$ is calculated via a cross-modal multi-head attention ($CrossMultiAttn$) as:
	\begin{equation}
	\bm{Q}_v = \bm{Z}^{l}_v\bm{W}^{l}_{q}, ~~ \bm{K}_t = \bm{Z}^{l}_t\bm{W}^l_{k}, ~~ \bm{V}_t = \bm{Z}^{l}_t\bm{W}_v^{l},
	\end{equation}
	\begin{equation}
	\bm{Z}_{t\rightarrow v}^l = CrossMultiAttn(\bm{K}_t, \bm{Q}_v, \bm{V}_t),
	\end{equation}
	where $\bm{W}_q^{l}$, $\bm{W}^l_{k}$ and $\bm{W}_v\in\mathbb{R}^{d\times d}$ are model weights. After obtaining the $\bm{Z}_{t\rightarrow v}^l$, we inject it into the visual feature as
	\begin{equation}
	\tilde{\bm{Z}}^{l}_v = LN(\bm{Z}_{t\rightarrow v}^l\cdot{\bm{W}'}_v^l + \bm{Z}^{l}_{v}),
	\end{equation}
	where ${\bm{W}'}_v^l\in\mathbb{R}^{d\times d}$ is the model weight. Similarly, the output of each visual encoder layer can be collected to form $\bm{C}_v=\{\tilde{\bm{Z}}_v^{1}, \cdots, \tilde{\bm{Z}}_v^{L}\}$, as like Figure~\ref{fig:methodology_c}.
	
	\subsubsection{Summary-Worthy Information from Summary}
	Since some vital summary concepts are only available in the video content rather than transcripts (e.g., the `\texttt{ping pong}' in Figure~\ref{fig:introduction}), we also expect the visual encoder could have the ability to identify the novel summary-worthy visual feature that might not appear in transcripts. We are thus motivated to bridge the visual feature learning and the generated summary in a more direct manner, thus further enhancing our visual encoder.
	
	One potential way of benefiting our visual encoder is to distil the task-specific knowledge from other large-scale summarization models. However, most of the available ones are text-only methods, of which the knowledge is not suitable for our visual encoder. To this end, instead of the traditional knowledge distillation, we take advantage of self distillation~\cite{Zhang2019BeYO}, which has been proven to be an effective way to distil knowledge within the network itself. The basic idea is that we take the output of the decoder as the pseudo summary (i.e., hints) to teach the learning process of the visual encoder.
	
	Technically, we apply the average pooling on the output vector sequence $(\bm{w}_1, \dots, \bm{w}_{n_w})$ from the last layer of the decoder to obtain $\bm{w}\in\mathbb{R}^d$, which can be regarded as the global representative of the generated pseudo summary. Then we distil this summary-worthy knowledge into our visual encoder layer via a self distillation mechanism (SDM) as
	\begin{equation} \label{equ:sd}
	\mathcal{L}_{Sd} = \lambda\cdot\Vert \bm{w}-Linear(avg(\tilde{\bm{Z}}^{l}_{v})) \Vert,
	\end{equation}
	where $\lambda$ is the hyper-parameters, $avg(\cdot)$ is the average pooling layer. As seen from \eqref{equ:sd}, this knowledge distillation works by decreasing the distance between the feature maps of the pseudo summary from the decoder and of the visual feature from the visual encoder. Since the feature maps are from two modalities with different dimensions, an extra linear projector is added to project the visual one to the same dimension as the other. Finally, the training loss $\mathcal{L}$ of our method is a sum of the objectives as:
	\begin{equation}
	\mathcal{L} = \mathcal{L}_{Abs} + \mathcal{L}_{Sd}.
	\end{equation}
	
	During the training, as the visual feature in our SDM can gradually fit the feature map of the decoder output in a global representative way, the inexplicit knowledge and novel concepts that have not appeared in transcripts are injected into learning the visual feature, and thus we achieve a new summary-worthy collection $\bm{C}_v=\{\hat{\bm{Z}}_v^{1}, \cdots, \hat{\bm{Z}}_v^{L}\}$, see Figure~\ref{fig:methodology_d}.
	
	\subsection{Implementation Details}
	The BART-base model is adopted as the backbone of our model, in which $L=6$ for both the encoder and decoder. For the introduced auxiliary visual encoder, we use a $6$-layer encoder with 8 attention heads and a 768 feed-forward dimension. Following previous work~\cite{yu2021vision}, we set the max length of the generated summary to be 64 tokens; the decoding process can be stopped early if an End-of-Sequence (EOS) token is emitted. 
	The Adam~\cite{Kingma2014AdamAM} with $\beta_1=0.9$, $\beta_2=0.999$, and a weight decay of $1e^{-5}$ is employed as the optimizer. 
	
	\section{Experiments}
	
	\begin{table}[!t]
		\centering
		\begin{tabular}{l|ccc}
			\toprule
			Dataset & Train & Dev & Test \\ 
			\midrule
			How2 & 68336 & 2520 & 2127 \\
			How2-300 & 13167 & 150 & 127 \\
			MM-AVS & 836 & 104 & 105 \\
			\bottomrule
		\end{tabular}%
		\caption{The statistics of the three datasets.}
		\label{table:dataset}
	\end{table}
	
	\begin{table*}[!t]
		\centering
		\small
		\resizebox{0.8\linewidth}{!}{
			\begin{tabular}{lcccccccccc}
				\toprule
				\textbf{Method} & \textbf{R-1} & \textbf{R-2} & \textbf{R-L} & \textbf{B-1} & \textbf{B-2} & \textbf{B-3} & \textbf{B-4} & \textbf{M} & \textbf{C} & \textbf{CF}\\
				\midrule
				\multicolumn{11}{c}{\textit{Transcript}}\\
				\midrule
				S2S$^\dagger$ & 58.6 & 40.6 & 53.8 & 55.2 & 45.6 & 39.9 & 35.8 & 27.6 & 2.35 & -\\
				PG$^\dagger$ & 57.2 & 39.5 & 52.8 & 55.3 & 45.6 & 39.8 & 35.7 & 26.8 & 2.13 & -\\
				TF$^\dagger$ & 59.0 & 41.0 & 54.3 & 56.6 & 46.7 & 40.8 & 36.6 & 27.7 & 2.30 & -\\
				T5$^*$ & 62.8 & 45.0 & 57.5 & 60.5 & 50.4 & 44.2 & 39.6 & 30.6 & 2.76 & 61.7\\
				BART$^*$ & 64.0 & 46.4 & 58.9 & 62.4 & 52.6 & 46.4 & 42.0 & 31.7 & 2.97 & 63.9\\
				\midrule
				\multicolumn{11}{c}{\textit{Transcript + Video}}\\
				\midrule
				HA (RNN)$^\dagger$ & 60.3 & 42.5 & 55.7 & 57.2 & 47.7 & 41.8 & 37.5 & 28.8 & 2.48 & -\\
				HA (TF)$^\dagger$ & 60.2 & 43.1 & 55.9 & 58.6 & 48.3 & 43.3 & 38.1 & 28.9 & 2.51 & -\\
				MFFG (RNN)$^\dagger$ & 62.3 & 46.1 & 58.2 & 59.1 & 50.4 & 45.1 & 41.1 & 30.1 & 2.69 & -\\
				MFFG (TF)$^\dagger$ & 61.6 & 45.1 & 57.4 & 60.0 & 50.9 & 45.3 & 41.3 & 29.9 & 2.67 & -\\
				SFFG$^\ddagger$ & 63.2 & 46.4 & 58.9 & 61.5 & 52.3 & 46.5 & 42.4 & 31.6 & 2.74 & -\\
				VG-T5$^*$ & 63.3 & 45.3 & 58.0 & 60.7 & 50.8 & 44.7 & 40.2 & 31.0 & 2.86 & 62.8\\
				VG-BART$^*$ & 68.0 & 51.4 & 63.3 & 65.2 & 56.3 & 50.4 & 46.0 & 34.0 & 3.28 & 69.7\\
				\midrule
				\multicolumn{11}{c}{\textit{Our Framework and the Variants}}\\
				\midrule
				$SWVR$ & \textbf{69.1} & \textbf{53.1} & \textbf{64.4} & \textbf{68.9} & \textbf{60.0} & \textbf{54.3} & \textbf{50.0} & \textbf{36.8} & \textbf{3.58} & \textbf{72.8}\\
				~~- w/o D & 68.7 & 52.6 & 63.9 & 68.5 & 59.5 & 53.7 & 49.3 & 36.5 & 3.42 & 72.3\\
				~~- w/o D+B & 68.2 & 51.5 & 63.4 & 66.5 & 57.3 & 51.5 & 47.2 & 35.9 & 3.30 & 71.9 \\
				~~- w/o D+B+A & 67.0 & 50.5 & 61.8 & 64.7 & 55.6 & 49.8 & 45.4 & 33.3 & 3.24 & 71.5  \\
				\bottomrule
			\end{tabular}
		}
		\caption{Evaluation results of baselines and our proposed models on the How2 dataset, where R, B, M, C, and CF stand for ROUGE, BLEU, MENTOR, CIDEr, and Content F1, respectively. Results with $^\dagger$, $^\ddagger$, and $^*$ marks are taken from \protect\cite{Liu2020MultistageFW}, \protect\cite{Liu2022AbstractiveSF}, and \protect\cite{yu2021vision}, respectively. Abbreviations D, B, and A stand for self distillation, bi-directional visual-language attention, and auxiliary visual encoder modules, respectively.}
		\label{table:how2}
	\end{table*}
	
	\begin{table}[!t]
		\centering
		\small
		\resizebox{0.75\linewidth}{!}{
			\begin{tabular}{lccc}
				\toprule
				\textbf{Method} & \textbf{R-1} & \textbf{R-2} & \textbf{R-L}\\
				\midrule
				\multicolumn{4}{c}{\textit{Transcript}}\\
				\midrule
				S2S$^\ddagger$ & 46.01 & 25.16 & 39.98 \\
				T5 & 55.36 & 36.01 & 49.73\\
				BART & 56.56 & 37.22 & 50.44\\
				\midrule
				\multicolumn{4}{c}{\textit{Transcript + Video}}\\
				\midrule
				MFFG (RNN)$^\ddagger$ & 48.53 & 28.69 & 44.08\\
				MFFG (TF)$^\ddagger$ & 49.27 & 28.26 & 43.41\\
				SFFG$^\ddagger$ & 50.60 & 30.38 & 44.67 \\
				VG-T5 & 57.72 & 36.80 & 50.37\\
				VG-BART & 58.24 & 37.99 & 51.46\\
				\midrule
				Ours & \textbf{59.92} & \textbf{40.75} & \textbf{53.81}\\
				\bottomrule
			\end{tabular}
		}
		\caption{Evaluation results of baselines and our proposed models on the How2-300 dataset. Results with $^\ddagger$ mark are taken from \protect\cite{Liu2022AbstractiveSF}.}
		\label{table:how2-300}
	\end{table}
	
	\begin{table}[!t]
		\centering
		\small
		\resizebox{0.75\linewidth}{!}{
			\begin{tabular}{lccc}
				\toprule
				\textbf{Method} & \textbf{R-1} & \textbf{R-2} & \textbf{R-L}\\
				\midrule
				\multicolumn{4}{c}{\textit{Without Training on the How2 dataset}}\\
				\midrule
				T5 & 22.36 & 9.00 & 17.96\\
				BART & 22.97 & 9.02 & 17.65\\
				VG-T5 & 12.96 & 1.12 & 10.82\\
				VG-BART & 13.95 & 1.20 & 11.59\\
				Ours & 23.67 & 10.23 & 18.97\\
				\midrule
				\multicolumn{4}{c}{\textit{First Training on the How2 dataset}}\\
				\midrule
				T5 & 23.46 & 9.95 & 18.\\
				BART & 23.36 & 9.56 & 17.97\\
				VG-T5 & 25.11 & 10.50 & 19.80\\
				VG-BART & 25.04 & 11.00 & 20.06\\
				Ours & \textbf{26.08} & \textbf{12.46} & \textbf{21.31}\\
				\bottomrule
			\end{tabular}
		}
		\caption{Evaluation results of baselines and our proposed models on the MM-AVS dataset.}
		\label{table:mm-avs}
	\end{table}
	
	\subsection{Experimental Setups}
	\paragraph{Datasets and Evaluation Metrics:}
	We evaluate the proposed $SWVR$ on three public datasets, including How2, How2-300~\cite{sanabria2018how2}, and MM-AVS~\cite{Fu2021MMAVSAF} dataset. The statistic of datasets is shown in Table~\ref{table:dataset}. We follow \cite{yu2021vision} to adopt the ROUGE-\{1,2,L\}~\cite{Lin2003AutomaticEO}, BLEU-\{1,2,3,4\}~\cite{Papineni2002BleuAM}, METEOR~\cite{Denkowski2011Meteor1A}, CIDEr~\cite{Vedantam2014CIDErCI}, and Content F1~\cite{Palaskar2019MultimodalAS} as the evaluation criteria. 
	
	\paragraph{Bacelines:}
	We compare $SWVR$ with the following two groups of baselines. \emph{1) Methods using transcript only:} In this group of baselines, we pick the existing text-only summarization methods, including S2S~\cite{luong2015effective}, PG~\cite{see2017get}, TF~\cite{vaswani2017attention}, T5~\cite{raffel2020exploring}, and BART~\cite{lewis2020bart}. \emph{2) Methods using both transcript and video:} We consider strong MAS baselines including HA(RNN/Transformer)~\cite{Palaskar2019MultimodalAS}, MFFG(RNN/Transformer)~\cite{Liu2020MultistageFW}, VG-GPLM(T5/BART)~\cite{yu2021vision}, and SFFG~\cite{Liu2022AbstractiveSF}.
	
	\subsection{Experimental Results}
	The results of How2 datasets are shown in Table~\ref{table:how2}. As observed, our proposed model significantly outperforms all previous methods, and we take an average improvement of 5.72\% over all the criterion compared with the strongest baseline VG-BART. It confirms the effectiveness of our proposed summary-worthy mechanism. We can also see that the performance of abstractive summarization with the help of both transcripts and video content significantly outperform transcript-only summarization models, demonstrating that visual features contain valuable information that is complementary to the transcript.
	
	As for the How2-300 dataset, it can be observed from Table~\ref{table:how2-300} that our model still performs the best. An interesting found is that, compared with the VG-BART model, the average improvement 4.9\% (over R1, R2, and RL) on How2-300 is larger than that of 2.2\% in the How2 dataset, namely, our model can work even better on small datasets like How2-300. We conjecture that it is because the general-purpose visual feature extractors require more training data to distinguish the task-specific clues that are valuable for the MAS task, so they don't work very well on small datasets. In contrast, thanks to the superiority of our proposed BVLA and SDM, the key and novel information is easier to capture and align, therefore achieving much better performance on the smaller dataset.
	
	To further elaborate on the advantage of our model in the case of small datasets, we conduct experiments on MM-AVS, which has the minimum amount of data among benchmark datasets. We select T5, BART, VG-T5, and VG-BART for comparison. As shown in Table~\ref{table:mm-avs}, if we directly fine-tune the baselines on this dataset, severe over-fitting phenomena occurs in both the VG-T5 and VG-BART models. It can be further observed that the above two models can only coverage when first training on the larger How2 dataset, and then continuously fine-tuning on the smaller MM-AVS dataset. This can be attributed to the fact that it is hard for the baseline models to disentangle enough summary-worthy information from the general visual features if limited training data is provided. Instead, with the proposed summary-worthy mechanism, our model works surprisingly well on MM-AVS, no matter whether the model is first trained on the larger How2 dataset or not.
	
	\subsection{Ablation Study}
	\subsubsection{Impacts of Different Components}
	We evaluate the effectiveness of different components by gradually removing three modules, i.e., the SDM, BVLA, and auxiliary visual encoder. \emph{1) SWVR w/o D}: the self distillation is first removed; in such case, our model can only leverage the summary-related information from transcripts and becomes Fig.\ref{fig:methodology_c}. \emph{2) SWVR w/o D+B}: The BVLA is further replaced with the uni-directional VLA, then our model can be summarized as Fig.\ref{fig:methodology_b}. Note that in this case, our model is not aware of any summary-worthy information. \emph{3) SWVR w/o D+B+A}: Lastly, the introduced auxiliary visual encoder is dropped, and our model is degraded to the simplest form Fig.\ref{fig:methodology_a}. 
	
	\begin{figure}[ht]
		\centering
		\begin{minipage}[t]{0.48\linewidth}
			\centering
			\includegraphics[scale=0.3]{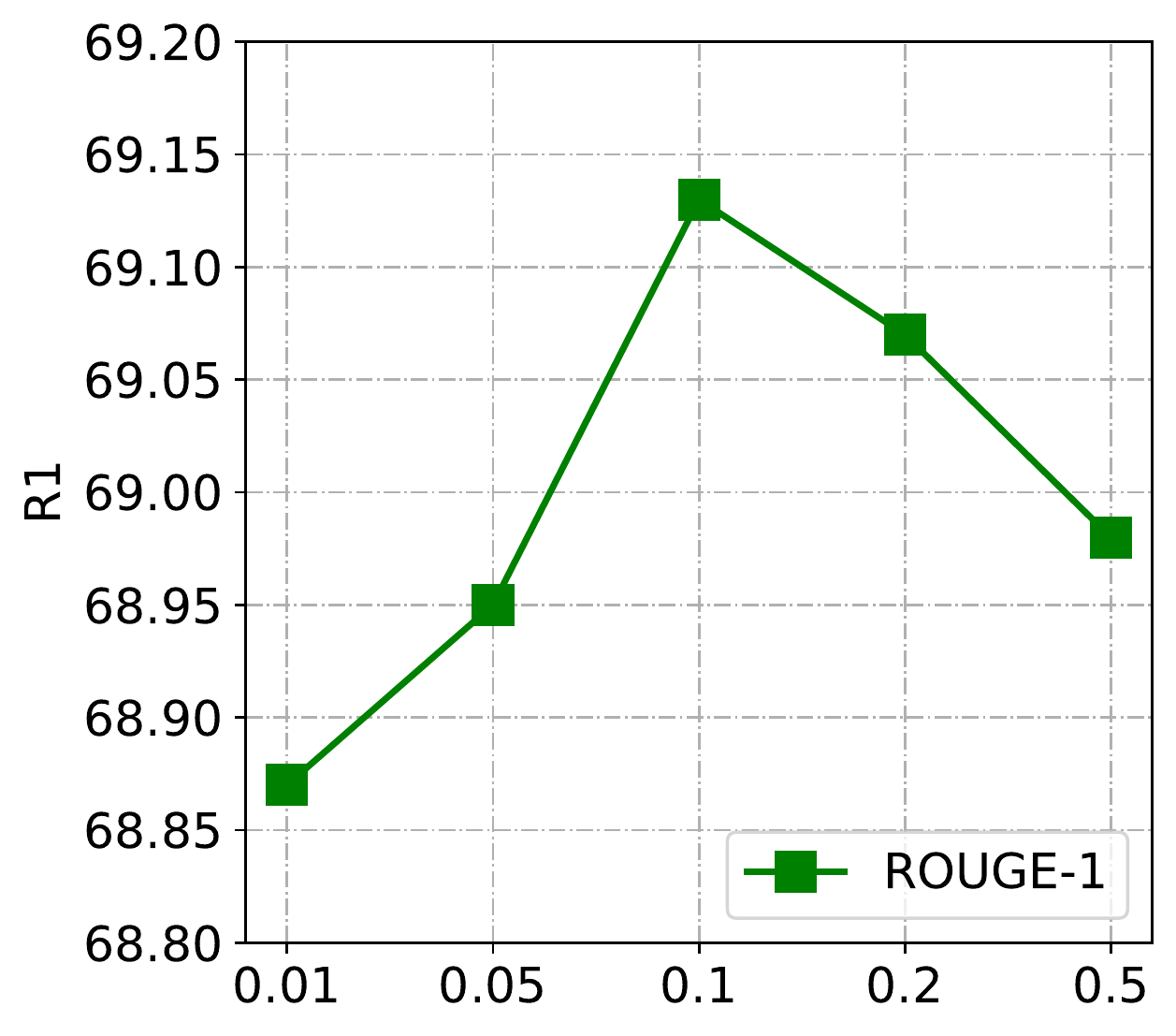}
		\end{minipage}
		\begin{minipage}[t]{0.48\linewidth}
			\centering
			\includegraphics[scale=0.3]{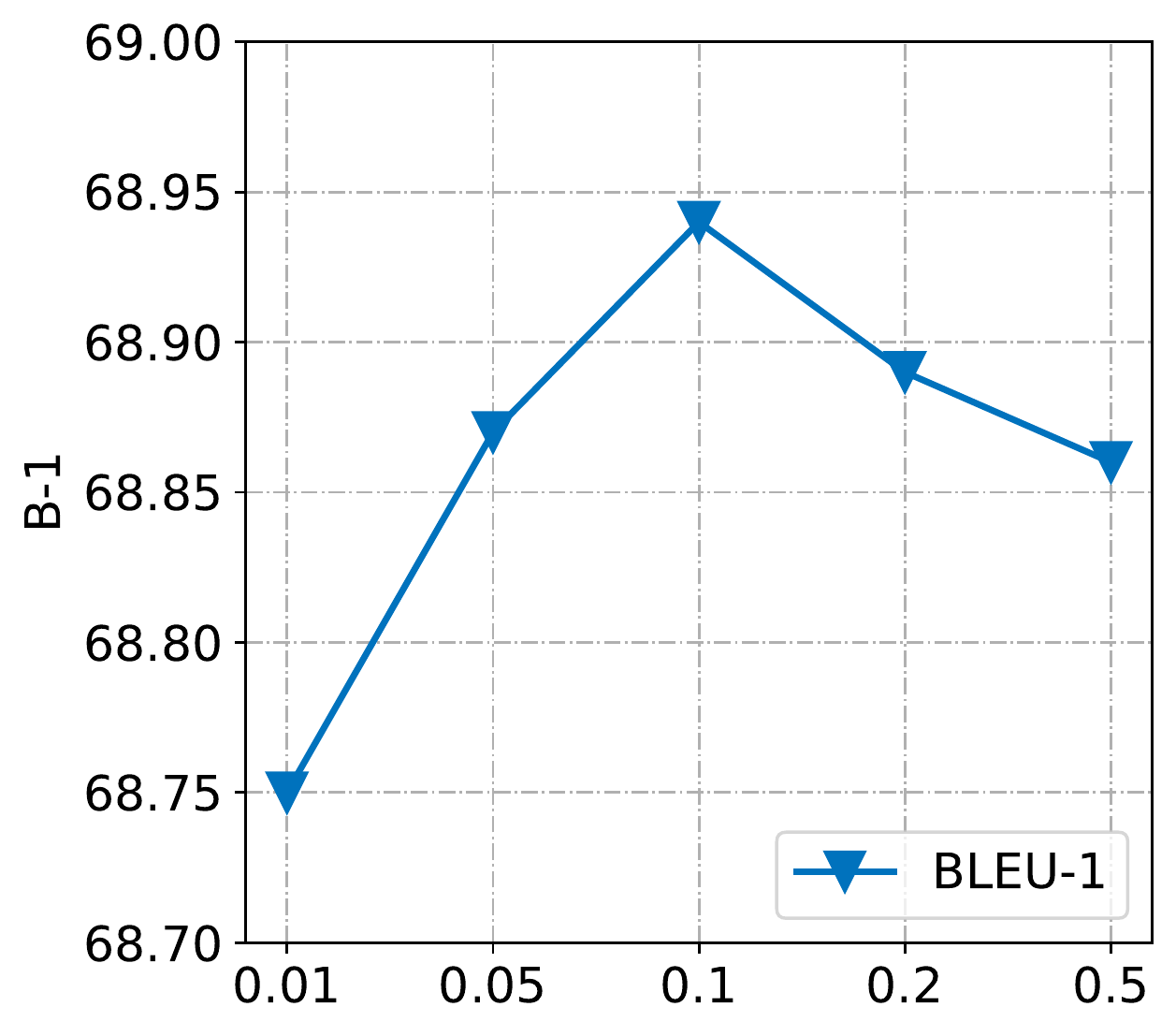}
		\end{minipage}
		\caption{Impacts of $\lambda$ on How2 dataset.}
		\label{fig:lambda}
	\end{figure}
	
	\begin{figure}[!t]
		\centering
		\includegraphics[scale=0.35]{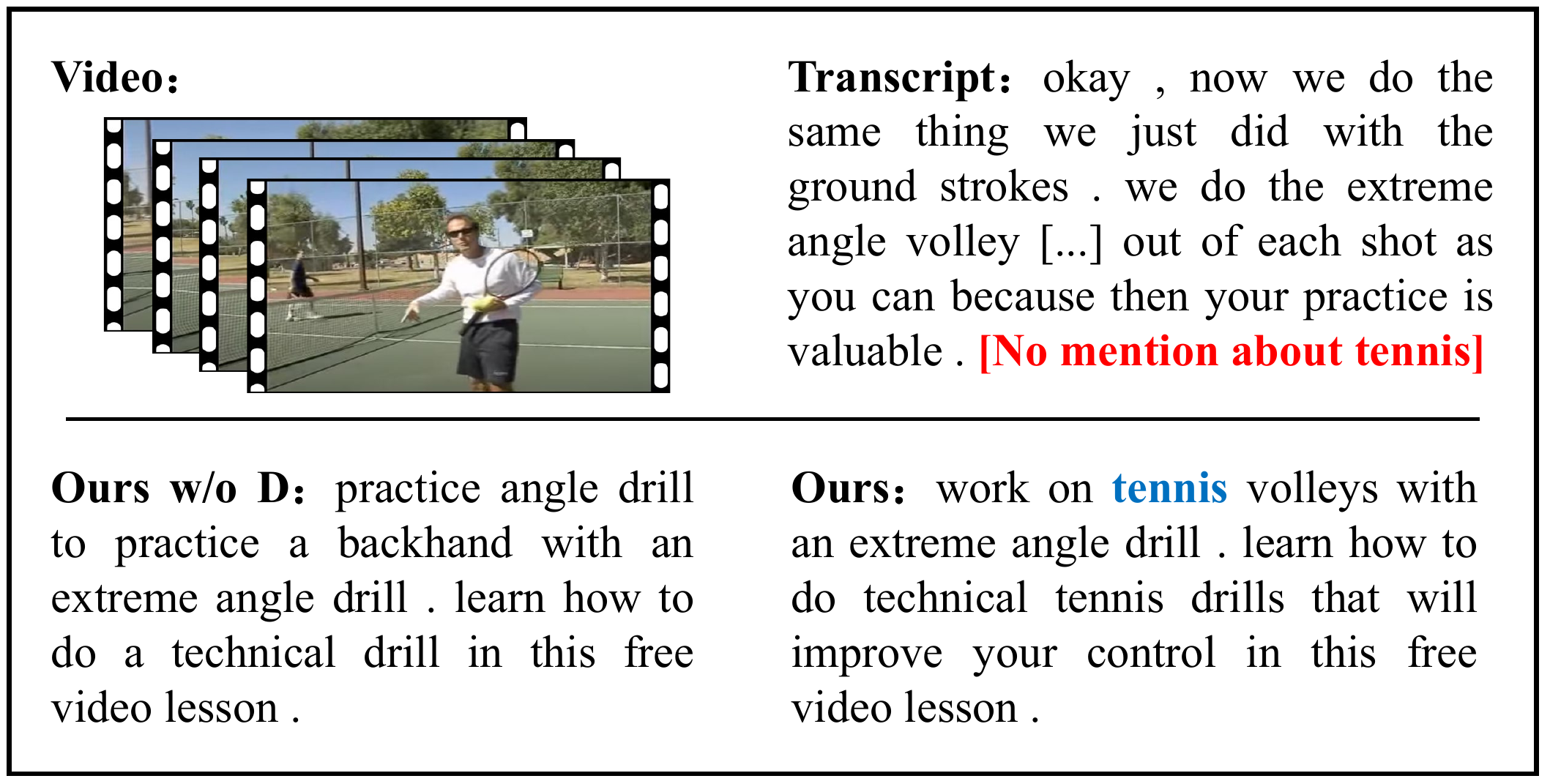}
		\caption{An example of case study. We show the generated summaries of model using summary-related and summary-worthy visual feature, respectively.}
		\label{fig:case_study}
	\end{figure}
	
	We compare the three variants with the complete $SWVR$ on the How2 dataset, and results are shown in Table~\ref{table:how2}. The general trend is that eliminating any of the modules (e.g., from SWVR w/o D to w/o D+B+A) will negatively impact the model's performance, confirming each component's benefits. Moreover, it is noticeable that the introduction of SMD can bring an improvement of 1.25\% on average, suggesting the importance of equipping our visual encoder with the ability to capture summary-worthy values from the pseudo summary.
	
	\subsubsection{Impacts of the Hyper-parameter of $\lambda$}
	In SDM, we introduce the hyper-parameter $\lambda$, which controls the trade-off between the cross-entropy loss and the self distillation loss. To analysis the sensitivity of $\lambda$, we manually select the values of $\lambda$ from \{0.01, 0.05, 0.1, 0.2, 0.5\}. ROUGE-1 and BLEU-1 w.r.t $\lambda$ on How2 datasets are illustrated in Figure~\ref{fig:lambda}. It can be observed that the performance of $SWVR$ first increases and gets the peak when $\lambda$ is not greater than 0.1. Afterward, the improvement is neutralized and even lost if $\lambda$ becomes larger. This may be attributed to the fact that a simple linear layer in \eqref{equ:sd} may not mitigate the difference between different modalities, and thus, we speculate that a more well-designed transformation may further discover the poentials of self distillation, which however, is beyond the scope of this paper.
	
	\subsection{Case Study}
	We conduct a case study to empirically exhibit the effectiveness of using summary-worthy visual information. To do so, we conduct experiments using two cases: one only exploits summary-worthy information from the transcript (see Fig.\ref{fig:methodology_c}), while the other exploits summary-worthy information from both the transcript and the generated pseudo summary (Fig.\ref{fig:methodology_d}).
	Data samples can be seen in Figure~\ref{fig:case_study}, the former case of only leveraging the summary-worthy values from the transcript fails to predict the `\texttt{tennis}', which concept is only available in video frames. With the assistance of distilled knowledge from the pseudo summary, our model captures the novel concept, or we call it visual object (i.e., \texttt{tennis}), from the video frames and successfully generates a high-quality summary result.
	
	\begin{table}[t]
		\centering
		\resizebox{0.9\linewidth}{!}{
			\begin{tabular}{cccccc|ccc}
				\toprule
				\multicolumn{6}{c}{\textbf{Adoption of BVLA}} & \multirow{2}{*}{\textbf{R-1}} & \multirow{2}{*}{\textbf{R-2}} & \multirow{2}{*}{\textbf{R-L}} \\ \cmidrule{1-6}
				1 & 2 & 3 & 4 & 5 & 6 & & &\\ \midrule
				\xmark & \xmark & \xmark & \xmark & \xmark & \xmark & 64.0 & 46.4 & 58.9\\ \midrule
				\cmark & \cmark & \cmark & \cmark & \cmark & \cmark & 65.7 & 49.3 & 60.7 \\
				\xmark & \cmark & \cmark & \cmark & \cmark & \cmark & 66.2 & 49.7 & 61.2 \\
				\xmark & \xmark & \cmark & \cmark & \cmark & \cmark & 67.3 & 51.1 & 62.3 \\
				\xmark & \xmark & \xmark & \cmark & \cmark & \cmark & \textbf{68.7} & \textbf{52.6} & \textbf{63.9} \\
				\xmark & \xmark & \xmark & \xmark & \cmark & \cmark & 68.4 & 52.0 & 63.5\\
				\xmark & \xmark & \xmark & \xmark & \xmark & \cmark & 67.9 & 51.5 & 62.8\\ \bottomrule
			\end{tabular}
		}
		\caption{How2 dataset performance of adopting BVLA at different locations in the encoder of \textit{SWVR w/o D} (i.e., the self distillation mechanism is removed). \cmark~ indicates the adoption at a certain layer and \xmark~ indicates non-adoption.}
		\label{tab:BVLA}
	\end{table}
	
	\begin{table}[t]
		\centering
		\resizebox{0.9\linewidth}{!}{
			\begin{tabular}{cccccc|ccc}
				\toprule
				\multicolumn{6}{c}{\textbf{Utilization of SDM}} & \multirow{2}{*}{\textbf{R-1}} & \multirow{2}{*}{\textbf{R-2}} & \multirow{2}{*}{\textbf{R-L}} \\ \cmidrule{1-6}
				1 & 2 & 3 & 4 & 5 & 6 & & &\\ \midrule
				\xmark & \xmark & \xmark & \xmark & \xmark & \xmark & 65.7 & 49.3 & 60.7\\ \midrule
				\cmark & \cmark & \cmark & \cmark & \cmark & \cmark & 66.1 & 49.7 & 60.9\\
				\xmark & \cmark & \cmark & \cmark & \cmark & \cmark & 66.5 & 50.0 & 61.4\\
				\xmark & \xmark & \cmark & \cmark & \cmark & \cmark & 67.0 & 50.4 & 61.7\\
				\xmark & \xmark & \xmark & \cmark & \cmark & \cmark & 67.3 & 50.9 & 62.1\\
				\xmark & \xmark & \xmark & \xmark & \cmark & \cmark & \textbf{67.8} & \textbf{51.4} & \textbf{62.5}\\
				\xmark & \xmark & \xmark & \xmark & \xmark & \cmark & 67.5 & 50.8 & 61.9\\ 
				\bottomrule
			\end{tabular}
		}
		\caption{How2 dataset performance of utilizing SDM at different layers of the visual encoder. In this case, we force the model to adopt BVLA at all the locations. \cmark~ indicates the utilization of SDM a certain layer while \xmark~ indicates non-utilization.}
		\label{tab:SDM}
	\end{table}
	
	\subsection{Where to Adopt the BVLA and SDM}
	In this section, we aim to further investigate the optimal location where to insert the BVLA and SDM modules. As depicted in Table~\ref{tab:BVLA}, in general, leveraging BVLA to inject the cross-modal features into the unimodal representation can significantly boost the model's performance. A similar phenomenon of the utilization of SDM can also be observed in Table~\ref{tab:SDM}. Furthermore, we observe that inserting the BVLA or SDM modules to the top layers can achieve the best performance. We speculate that the lower layers of the encoder may tend to capture the local and low-level semantics, which are usually modality-specific. This makes it challenging for the model to capture better cross-modal semantic interaction. In contrast, the top layers of the encoder may primarily exploit the global and high-level semantics, where the high-level abstract information is more accessible for the model to perform cross-modal interaction. This observation is also consistent with \cite{xu2021vlm}. Given this conclusion, our best model reported in Table~\ref{table:how2} adopts the BVLA and SDM in \{4,5,6\}-th and \{5,6\}-th encoder layers, respectively.
	
	\section{Conclusion}
	In this paper, we propose to learn summary-worthy visual features to boost the MAS task. We introduce a bi-directional visual-language attention (BVLA) mechanism and a self distillation mechanism (SDM) to encourage the visual encoder to exploit the summary-worthy information from the textual data or the knowledge that distills from the pseudo summary. In addition, we enumerate almost all possible ways to combine and evaluate BVLA and SDM modules in the visual encoder. Experiments results show that our proposed method significantly outperforms all strong baselines on three public datasets. Further analysis demonstrates each component's effectiveness and suggests that the higher layers of the visual encoder are the optimal places to employ the BVLA and SMD.
	
	\section*{Acknowledgments}
	We are grateful to ShuTing Cai for perfecting this article. This work is supported by the National Natural Science Foundation of China (No. 62276280, U1811264), Key R\&D Program of Guangdong Province (No. 2018B010107005), Natural Science Foundation of Guangdong Province (No. 2021A1515012299), Science and Technology Program of Guangzhou (No. 202102021205).
	
	\bibliographystyle{named}
	\bibliography{ijcai23}
\end{document}